\documentclass[twocolumn]{bmcart}

\usepackage{amsthm,amsmath}
\RequirePackage{natbib}
\RequirePackage{hyperref}
\usepackage[utf8]{inputenc} %unicode support

\usepackage{xr}
\usepackage{graphicx}
\usepackage{booktabs}
\usepackage{dsfont}
\usepackage{amsmath}
\usepackage{amssymb}
\usepackage{caption}
\usepackage{subcaption}
\newcommand{\R}{\mathbb R}
\newcommand{\bP}{\mathbb P}
\newcommand{\norm}[1]{\|#1\|}

\startlocaldefs
\endlocaldefs

\begin{document}

\begin{frontmatter}

\begin{fmbox}
\dochead{Research}

\title{Comparison of methods for early-readmission prediction in a high-dimensional heterogeneous covariates and time-to-event outcome framework}

\author[
   addressref={aff1},
   corref={aff1},
   email={simon.bussy@gmail.com}
]{\inits{SB}\fnm{Simon} \snm{Bussy}}
\author[
   addressref={aff2,aff3}        
]{\inits{RV}\fnm{Rapha\"el} \snm{Veil}}
\author[
   addressref={aff2,aff3}        
]{\inits{VL}\fnm{Vincent} \snm{Looten}}
\author[
   addressref={aff2,aff3}
]{\inits{AB}\fnm{Anita} \snm{Burgun}}
\author[
   addressref={aff1,aff4}        
]{\inits{SG}\fnm{St\'ephane} \snm{Ga\"iffas}}
\author[
   addressref={aff5}        
]{\inits{AG}\fnm{Agathe} \snm{Guilloux}}
\author[
   addressref={aff6,aff7}        
]{\inits{BR}\fnm{Brigitte} \snm{Ranque}}
\author[
   addressref={aff2,aff3}
]{\inits{ASJ}\fnm{Anne-Sophie} \snm{Jannot}}

\address[id=aff1]{
  \orgname{Laboratoire de Probabilités Statistique et Modélisation (LPSM), UMR 8001, Sorbonne University},
  \street{4 Place Jussieu},
  \postcode{75005}
  \city{Paris},
  \cny{France}
}
\address[id=aff2]{                   
  \orgname{Assistance Publique-Hôpitaux de Paris, Biomedical Informatics and Public Health Department, European Georges Pompidou Hospital}, 
  \street{20 Rue Leblanc},                     
  \postcode{75015}                                
  \city{Paris},                             
  \cny{France}                                    
}
\address[id=aff3]{
  \orgname{INSERM UMRS 1138, Eq22, Centre de Recherche des Cordeliers, Université Paris Descartes},
  \street{15 Rue de l'École de Médecine},
  \postcode{75006}
  \city{Paris},
  \cny{France}
}
\address[id=aff4]{
  \orgname{CMAP, UMR 7641 École Polytechnique CNRS},
  \street{Route de Saclay},
  \postcode{91128}
  \city{Palaiseau},
  \cny{France}
}
\address[id=aff5]{
  \orgname{LAMME, Univ Evry, CNRS, Universit\'e Paris-Saclay},
  \street{23 boulevard de France},
  \postcode{91025}
  \city{Evry},
  \cny{France}
}
\address[id=aff6]{
  \orgname{INSERM UMRS 970, Université Paris Descartes},
  \street{56 rue Leblanc},
  \postcode{75015}
  \city{Paris},
  \cny{France}
}
\address[id=aff7]{
  \orgname{Assistance Publique-Hôpitaux de Paris, Internal Medicine Department, Georges Pompidou European Hospital},
  \street{20 Rue Leblanc},
  \postcode{75015}
  \city{Paris},
  \cny{France}
}

\begin{abstractbox}

\begin{abstract}

\parttitle{Background}
Choosing the most performing method in terms of outcome prediction or variables selection is a recurring problem in prognosis studies, leading to many publications on methods comparison. But some aspects have received little attention. First, most comparison studies treat prediction performance and variable selection aspects separately. Second, methods are either compared within a binary outcome setting (based on an arbitrarily chosen delay) or within a survival setting, but not both. 
In this paper, we propose a comparison methodology to weight up those different settings both in terms of prediction and variables selection, while incorporating advanced machine learning strategies.

\parttitle{Methods}
Using a high-dimensional case study on a sickle-cell disease (SCD) cohort, we compare 8 statistical methods. In the binary outcome setting, we consider logistic regression (LR), support vector machine (SVM), random forest (RF), gradient boosting (GB) and neural network (NN); while on the survival analysis setting, we consider the Cox Proportional Hazards (PH), the CURE and the C-mix models. We then compare performances of all methods both in terms of risk prediction and variable selection, with a focus on the use of Elastic-Net regularization technique.

\parttitle{Results}
Among all assessed statistical methods assessed, the C-mix model yields the better performances in both the two considered settings, as well as interesting interpretation aspects.
There is some consistency in selected covariates across methods within a setting, but not much across the two settings. 

\parttitle{Conclusions}
It appears that learning withing the survival setting first, and then going back to a binary prediction using the survival estimates significantly enhance binary predictions.

\end{abstract}


\end{abstractbox}

\end{fmbox}

\end{frontmatter}

\section*{Background}

Recently, many statistical developments have been performed to tackle prognostic studies analysis. Beyond accurate risk estimation, interpretation of the results in terms of covariates importance is required to assess risk factors, with the ultimate aim of developing better diagnostic and therapeutic strategies~\cite{pittman2004integrated}.

In most studies, covariate selection ability and model prediction performance are regarded separately. On the one hand, a considerable amount of studies report on covariates relevancy in multivariate models, mostly in the form of ajusted odds ratio~\cite{little2009strengthening} (for instance using logistic regression (LR) model~\cite{bender1996logistic, mikolajczyk2008evaluation}) without reporting on the method's prediction performance (goodness-of-fit and overfitting aspects are neglected); namely disregarding the question: \textit{is the model prediction still accurate on new data, unseen during the training phase?}
While on the other hand, most studies focusing on a method's predictive performance do not mention its variable selection ability~\cite{guyon2003introduction}, thus making it not well suited for the high-dimensional setting. Such settings are becoming increasingly common in a context where the number of available covariates to consider as  potential risk factors is tremendous, especially with the development of electronic health record (EHR).

In this paper, we discuss both aspects (prediction performance and covariates selection) for all considered methods, with a particular emphasis on the \textit{Elastic-Net} regularization method~\cite{zou2005regularization}. Regularization has emerged as a dominant theme in machine learning and statistics. It provides an intuitive and principled tool for learning from high-dimensional data.

Then, a lot of studies consider prognosis as a binary outcome, namely whether the event-of-interest (death, relapse or hospital readmission for instance) occurs whithin a pre-specified period of time we denote $\epsilon$~\cite{tong2016comparison, rich1995multidisciplinary, vinson1990early, boulding2011relationship}. In the following, we refer to this framework as the \textit{binary outcome setting}, and we denote $T \geq 0$ the time elapsed before the event-of-interest and $X \in \R^d$ the vector of $d$ covariates recorded at the hospital during a stay. In this setting, we are interested in predicting $T\leq\epsilon$.
Such an \textit{a priori} choice for $\epsilon$ is questionable, since any conclusion regarding both prediction and covariates relevancy is completely conditioned on the threshold value $\epsilon$~\cite{chen2012assessment}. Hence, it is hazardous to make general inference on the probability distribution of the time-to-event outcome given the covariates from such a restrictive binary prediction setting. 

An alternative setting to model prognosis is the survival analysis one, that takes the quantitative censored times as outcomes.
The time $T$ is right censored since in practice, some patients have not been readmitted before the end of follow-up.
In the following, we refer to this setting as the \textit{survival analysis setting}~\cite{kleinbaum2010survival} and we denote $Y$ the right-censored duration, that is $Y=\min(T,C)$ with $C$ the time when the patient is lost to follow-up.
Few studies compare the survival analysis and binary outcome settings and none of them considers simultaneously the prediction and the variable selection aspects in a high dimensional setting. For instance in~\citet{chen2012assessment}, only the Cox Proportional Hazards (PH) model~\cite{Cox1972JRSS} is considered in the survival analysis setting and a dimentionality reduction phase (or screening) is performed prior to the models comparison, as it is often the case~\cite{dai2006dimension, boulesteix2009optimal}.

Our case study focuses on hospital readmission following vaso-occlusive crisis (VOC) for patients with sickle-cell disease (SCD). SCD is the most frequent monogenic disorder worldwide. It is responsible for repeated VOC, which are acute painful episodes, utlimately resulting in increased morbidity and mortality~\cite{bunn1997pathogenesis, platt1991pain}. Although there are some studies regarding risk factors of early complications, only a few of them specifically addressed the question of early-readmission prediction after a VOC episode~\cite{brousseau2010acute, rees2003guidelines}.

For a few decades, hospital readmissions have been known to be responsible for huge costs~\cite{friedman2004rate, kocher2011hospital}; they are also a measure of health care quality. 
Today, hospitals have limited ressources they can allocate to each patient. Therefore, identifying patients at high risk of readmissions is a paramount question and predictive models are often used to tackle it. 

The purpose of this manuscript is to compare different statistical methods to analyse readmission. 
To make such comparisons, we consider both the predictive performance and the covariates selection aspect of each model, on the same high-dimensional set of covariates.

In the binary outcome setting, we consider LR~\cite{hosmer2013applied} and support vector machine (SVM)~\cite{scholkopf2002learning} with linear kernel, being both penalized with the Elastic-Net regularization~\cite{zou2005regularization} to deal with the high dimensional setting and avoid overfitting~\cite{hawkins2004problem}. We also consider random forest (RF)~\cite{breiman2001random}, gradient boosting (GB)~\cite{friedman2002stochastic} and artificial neural networks (NN)~\cite{yegnanarayana2009artificial}.

We then abstain from the \textit{a priori} threshold choice and consider the survival analysis setting. We apply first the Cox PH model~\cite{Cox1972JRSS}. We also apply the CURE model~\cite{farewell1982use, kuk1992mixture}, that considers one fraction of the population as cured or not subject to any risk of readmimssion. 
Finally, we consider the recently developped high dimensional C-mix mixture model~\cite{bussy2016}. The three considered models in this setting are also penalized with the Elastic-Net regularization. 

\section*{Methods}

\subsection*{Motivating case study}

We consider a monocentric retrospective cohort study of $n=286$ patients. George Pompidou University Hospital (GPUH) is an expertise center for SCD adult patients~\cite{bndmr}. Data is extracted from the GPUH Clinical Data Warehouse (CDW) using the i2b2 star-shaped standard~\cite{zapletal2010methodology}. 
It contains routine care data divided into several categories among them demographics, vital signs, diagnoses (ICD-10~\cite{world2004international}), procedures (French CCAM classification~\cite{trombert2003development}), EHR clinical data from structured questionnaires, free text reports, Logical Observation Identifiers Names and Codes (LOINC), biological test results, and Computerized Provider Order Entry (CPOE) drug prescriptions.
The sample included all stays from patients admitted to the internal medicine department for VOC (ICD-10 57.0 or 57.2) between January 1st 2010 and December 31st 2015.

Over half of the patients has only one stay during the follow-up period. 
We hence randomly sample one stay per patient and focus on the early-readmission risk afterwards. This enables us, in addition, to work on the \textit{independent and identically distributed} standard statistical framework.

\subsection*{Covariates}

We extracted demographic data (e.g. sex, date of birth, last known vital status), as well as both qualitative (e.g. the admission at any point during the stay to an ICU, the type of opioid drug received) and quantitative time-dependent variables (e.g. biological results, vital sign values, intraveinous opiod syringes parameters) regarding each stay. 

We also extracted all the free text reports from the patients’ EHR regardless of the source department and the stay. In order to facilitate variable extraction from such textual reports, we used a locally developed browser-accessible tool called FASTVISU~\cite{escudie2015reviewing}. This software is linked with the CDW, and allowed us to quickly check throughout these textual reports for highlighted information and to vote for variable status (e.g. SCD genotype) or value (e.g. baseline hemoglobinemia). Keywords using regular expressions are used to focus on specific mentions within the text. Variables extracted using this tool were the following: SCD genotype, baseline hemoglobinemia, medical history (with a focus on previous VOC complications and SCD-related chronic organ damages), and lifestyle related information. For time-dependent variables, status was determined per stay, including the ones that were not related to a VOC episode (e.g. annual check-ups).

We extracted for the included patients all stays encoded as VOC to derive time length from and until the respectively previous and consecutive stays.
Regarding demographic data, we derived the patient’s age at admission for each stay.
For each time-dependent covariate, all patient relative time series have different number of points and different length. We then propose a method to extract several covariates from each time series, to make the use of usual machine learning algorithms possible:
\begin{itemize}
  \item Regarding all vital parameters and oxygen use, we derived them by calculating the average value and the linear regression’s slope for the last 48 hours of the stay, as well as the delay between the end of oxygen support and the hospital discharge.
  \item Regarding biological variables, we only kept the ones that were measured at least once for more than 50\% of the stays. We considered the last measured value for each of them before discharge. Additionally, for covariates with at least 2 distinct measurements per stay, we considered the linear regression's slope for the last 48 hours of the stay. In order to maximize the amount of biological data, we also retrieved the biological values measured in the emergency department, prior to the administrative admission of the patient.
  \item For each time-dependent covariate and for each stay, we fit a distinct Gaussian process on the last 48 hours of the stay for all patient with at least 3 distinct measurements during this period, and extract the corresponding hyper-parameters as covariates for our problem.
\end{itemize}

Indeed, Gaussian processes are known to fit EHR data well; see for instance~\citet{pimentel2013modelling}, where a distinct Gaussian process is also fitted for each patient and each time-dependent covariate, in order to cluster patients into groups in the hyper-parameter space. In our study, we instead use the hyper-parameters as covariates in a supervised learning way.
We use Gaussian process with linear average function and a sum-kernel composed by a constant kernel which modifies the mean of the Gaussian process, a radial-basis function kernel, and a white kernel to explain the noise-component of the signal.

After a binary encoding of the categorical covariates, the final dimension of the working space (number of considered covariates) is $d=174$. Therefore, the number of patients is less than 2 times as many as the number of covariates, making it difficult to use standard regression techniques.
More details on data extraction, missing data imputation, as well as a precise list of all considered covariates, are given in Sections~\ref{sec:cov_creation}, ~\ref{sec:missing_data} and~\ref{sec:covariates_list} of Supplementary Material respectively.

\subsection*{Statistical methods and analytical strategies}

\subsubsection*{Binary outcome setting}

\begin{figure}[!htb]
\captionsetup{width=0.85\linewidth}
\centering
\includegraphics[width=.95\linewidth]{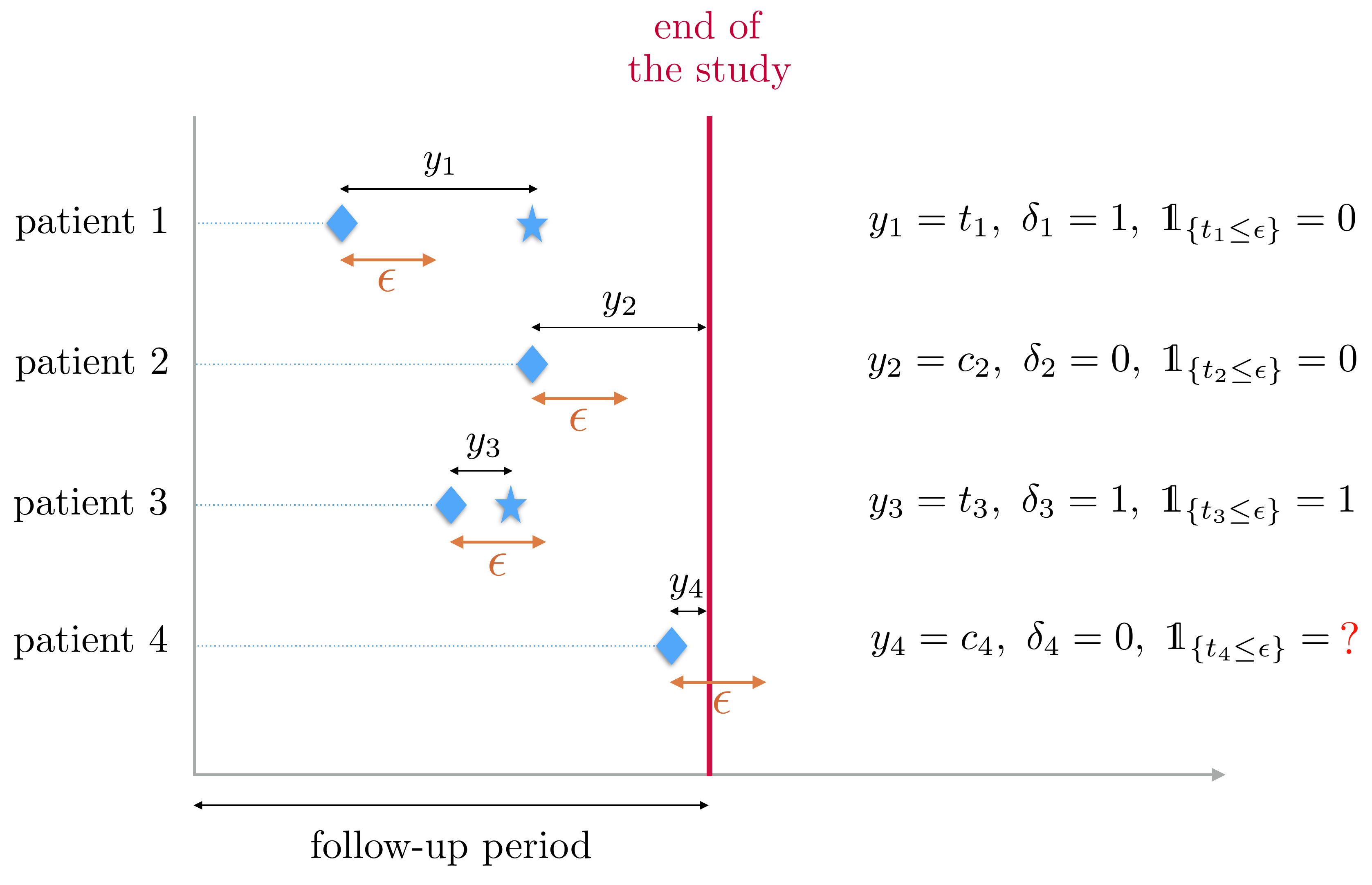}
\caption{Illustration of the problem of censored data that cannot be labeled when using a threshold $\epsilon$. $\delta_i = \mathds{1}_{\{T_i \leq C_i\}}$ is the censoring indicator which is equal to 1 if $Y_i$ is censored and 0 otherwise. In the binary outcome setting, patient 4 would be excluded.}
\label{fig:illustration_pb_binary}
\end{figure}

In this setting, we consider as early-readmission any readmission occuring within 30 days of hospital discharge after a previous hospital stay for VOC, the 30 days threshold being a standard choice in SCD studies~\cite{brousseau2010acute, frei2009risk}.
A first drawback of this setting (which is rarely mentionned) is that patients having both a censored time and $c_i \leq \epsilon$ have to be excluded from the procedure, since we do not know if $t_i \leq \epsilon$ or not. Figure~\ref{fig:illustration_pb_binary} gives an illustration of this last point.
In our case, 7 patients have to be excluded because of this issue.

We first consider LR~\cite{hosmer2013applied} and linear kernel SVM~\cite{scholkopf2002learning}, both penalized with the Elastic-Net regularization~\cite{zou2005regularization}. 
For a given model, using this penalization means adding the following term 
$\gamma \big( (1-\eta)\norm{\beta}_1 + (\eta/2) \norm{\beta}_2^2 \big)$ to the cost function (the negative likelihood for instance) in order to minimize it in $\beta \in \R^d$, a vector of coefficients that quantifies the impact of each biomedical covariates on the associated prediction task. This means that the Elastic-Net regularization term is a linear combination of the lasso ($\ell_1$) and ridge (squared $\ell_2$) penalties for a fixed $\eta \in (0,1)$, tuning parameter $\gamma$, and where we denote $\norm{\beta}_p = \big( \sum_{i=1}^d |\beta_i|^p \big)^{1 / p}$ the $\ell_p$-norm of $\beta$. One advantage of this regularization method is its ability to perform model selection (for the lasso part) and to pinpoint the most important covariates relatively to the prediction objective. On the other hand, the ridge part allows to handle potential correlation between covariates \citep{zou2005regularization}. The penalization parameter $\gamma$ is carefully chosen using the same cross-validation procedure~\cite{kohavi1995study} for all competing models. Note that in practice, the intercept is not regularized.

We also consider other machine learning algorithms in the ensemble methods class such as RF~\cite{breiman2001random} and GB~\cite{friedman2002stochastic}. For both algorithms, all hyper-parameters are tuned using a randomized search cross-validation procedure~\cite{bergstra2012random}. For instance for RF: the number of trees in the forest, the maximum depth of the tree or the minimum number of samples required to split an internal node.
Note also that regarding the covariates importance for RF and GB, we use the Gini importance~\cite{menze2009comparison}, defined as the total decrease in node impurity weighted by the probability of reaching that node (which is approximated by the proportion of samples reaching that node) averaged over all trees of the ensemble. That is why the corresponding coefficients are all positive for those two models, which is to be kept in mind.
Finally, we consider NN~\cite{yegnanarayana2009artificial} in the form of a multilayer perceptron neural network with one hidden layer. We use stochastic gradient-based optimizer for NN and rectified linear units activation function to get sparse activation and be able to compare covariate importance~\cite{glorot2011deep}. The regularization term as well as the number of neurons in the hidden layer are also cross-validated though a random search optimization.
Note that many studies in the literature choose hyper-parameters of the models, without mentioning any statistical procedure to determine them without \textit{a priori}~\cite{puddu2012artificial}. 

For all considered models in this setting, we use the reference implementations from the \texttt{scikit-learn} library~\citep{pedregosa2011scikit}.

\subsubsection*{Survival analysis setting}

The Cox PH model is by far the most widely used in the survival analysis setting; see \citet{Cox1972JRSS} and \citet{simon2011regularization} for the penalized version.
It is a regression model that describes the relation between intensity of events and covariates, given by $\lambda(t) = \lambda_0(t) \text{exp}(x^\top \beta)$
where $\lambda_0$ is a baseline intensity describing how the event hazard changes over time at baseline levels of covariates, and $\beta$ is a vector quantifying the multiplicative impact on the hazard ratio of each covariate. We use the \texttt{R} packages \texttt{survival} and \texttt{glmnet} to train this model.
An alternative to the Cox PH model is the CURE model~\cite{farewell1982use} that considers one fraction of the population as not subject to any risk of readmission, with a logistic function for the incidence part and a parametric survival model. We add an Elastic-Net regularization term and use the appropriate implementation detailed in Additional file 2.
Finally, we apply the C-mix model~\citep{bussy2016} that is designed to learn risk groups in a high dimensional survival setting. For a given patient $i$, it provides a marker $\pi_{\hat \beta}(x_i)$ estimating the probability that the patient is at high risk of early-readmission. Note that $\hat \beta$ denotes the estimate vector after the training phase for any model.

We randomly split data into a training set and a test set (30\% for testing, cross-validation is done on the training).
In both binary outcome and survival analysis settings, all the prediction performances are evaluated on the test set after the training phase, using the relevant metrics detailed hereafter. Note also that for all considered models (except RF and GB), continuous covariates are standardized through a preprocessing step, which allows proper comparability between the covariates' effects whithin each model.

\subsection*{Metrics used for analysis}
In the binary outcome setting, the natural metric used to evaluate performances is the
AUC~\citep{bradley1997use}. In the survival analysis setting, the natural equivalent is the C-index (implemented in the \texttt{python} package \texttt{lifelines}), that is $\bP[M_i > M_j | Y_i < Y_j , Y_i < \tau]$ with $i \neq j$ two independent patients, $\tau$ corresponding to the follow-up period duration~\cite{heagerty2005survival}, and $M_i$ the natural risk marker of the model for patient $i$: $\exp(x_i^\top \hat \beta)$ for the Cox PH model, the probability of being uncured for the CURE model and $\pi_{\hat \beta}(x_i)$ for the C-mix.

To compare the two settings, one can predict the survival function $\hat S_i$ for each model and for patients $i$ in the test set.
Then, for a given threshold $\epsilon$, one can now use $\hat S_i(\epsilon|X_i=x_i)$ for each model to predict whether or not $T_i \leq \epsilon$ on the test set -- relaying to the binary outcome setting -- thus assessing performances using the classical AUC score. Then, with $\epsilon = 30$ days, one can directly compare prediction performances with those obtained in the binary outcome setting. Details on the survival function estimation for each model are given in Section~\ref{sec:survival_estimation} of Supplementary Material.

Finally, we compute the pairwise Pearson correlation between the absolute (because of the positive vectors for RF and GB) covariates importance vectors of each method to obtain a similarity measure in terms of covariates selection~\citep{kalousis2007stability}.

\section*{Results}
Table~\ref{table:results} compares the prediction performances of the different methods in both considered settings using appropriate metrics. 
Corresponding hyper-parameters obtained by cross-validation are detailed in Section~\ref{sec:hyper_parameters} of Supplementary Material. 
\begin{table}[!h]
\centering
\caption{Comparison of prediction performances in the two considered settings, with best results in bold.}
\begin{tabular}{cccc}
\toprule
Setting & Metric  & Model & Score \\
\midrule
& & CURE & 0.718 \\
Survival analysis & C-index & Cox PH & 0.725 \\
& & C-mix & \textbf{0.754} \\
\midrule
& & SVM & 0.524 \\
& & GB & 0.561 \\
& & LR & 0.616 \\
& & NN & 0.707 \\
Binary outcome & AUC & RF & 0.738 \\
& & CURE ($\epsilon = 30$) & 0.831 \\
& & Cox PH ($\epsilon = 30$) & 0.855 \\
& & C-mix ($\epsilon = 30$) & \textbf{0.940} \\
\bottomrule
\end{tabular}
\label{table:results}
\end{table}

\begin{figure}[!htb]
\captionsetup{width=0.85\linewidth}
\centering
\includegraphics[width=.95\linewidth]{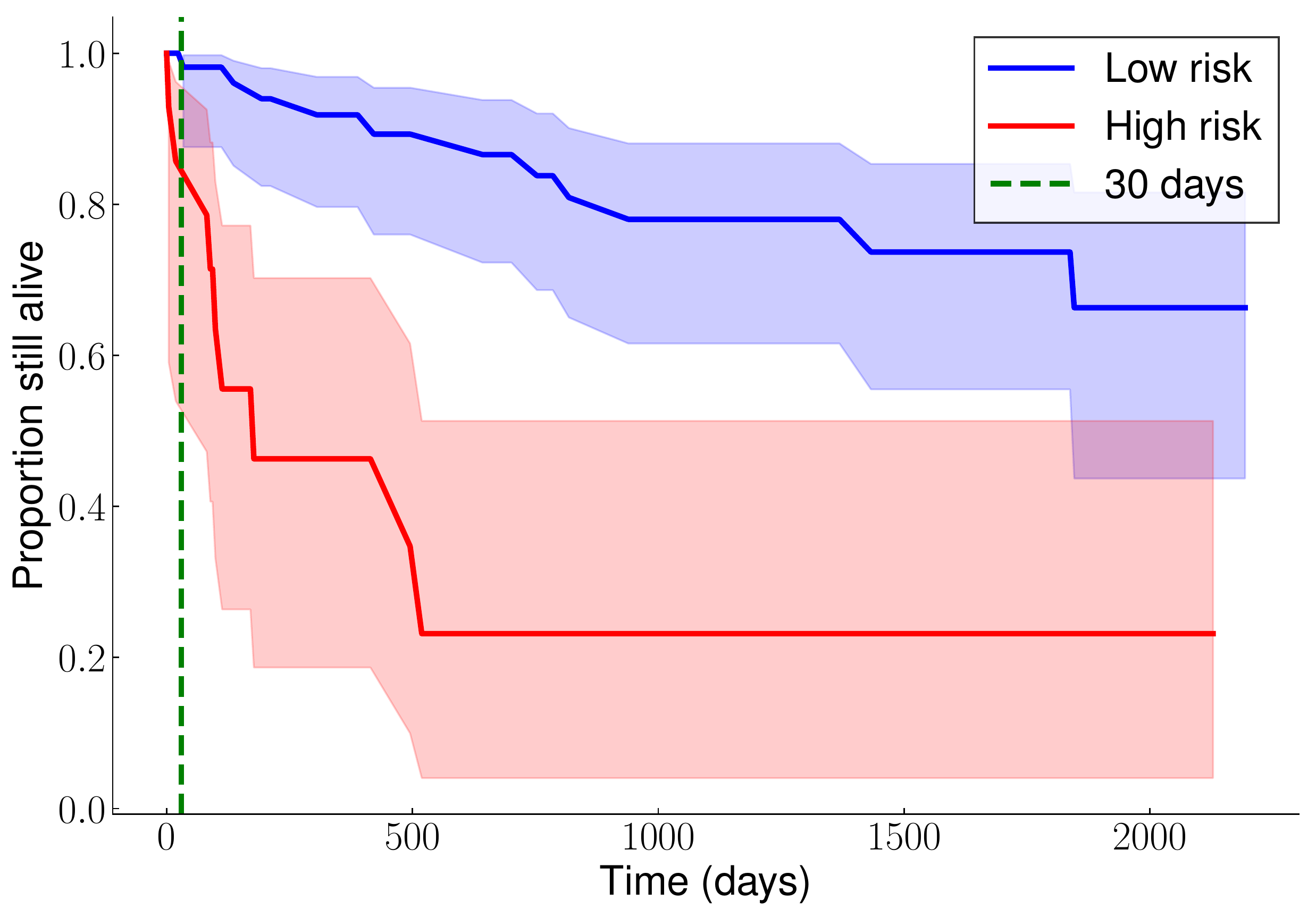}
\caption{Estimated survival curves per subgroups (blue for low risk and red for high risk) with the corresponding 95 \% confidence bands}
\label{fig:survival_curves}
\end{figure}

Thus, making binary predictions from survival analysis models using estimated survival function highly improves performances.
The C-mix yields the best results. Figure~\ref{fig:survival_curves} displays the estimated survival curves for the low and high risk of early-readmission subgroups learned by this model. Note the clear separation between the two subgroups.

\begin{figure}[!htb]
\captionsetup{width=0.85\linewidth}
\centering
\includegraphics[width=.95\linewidth]{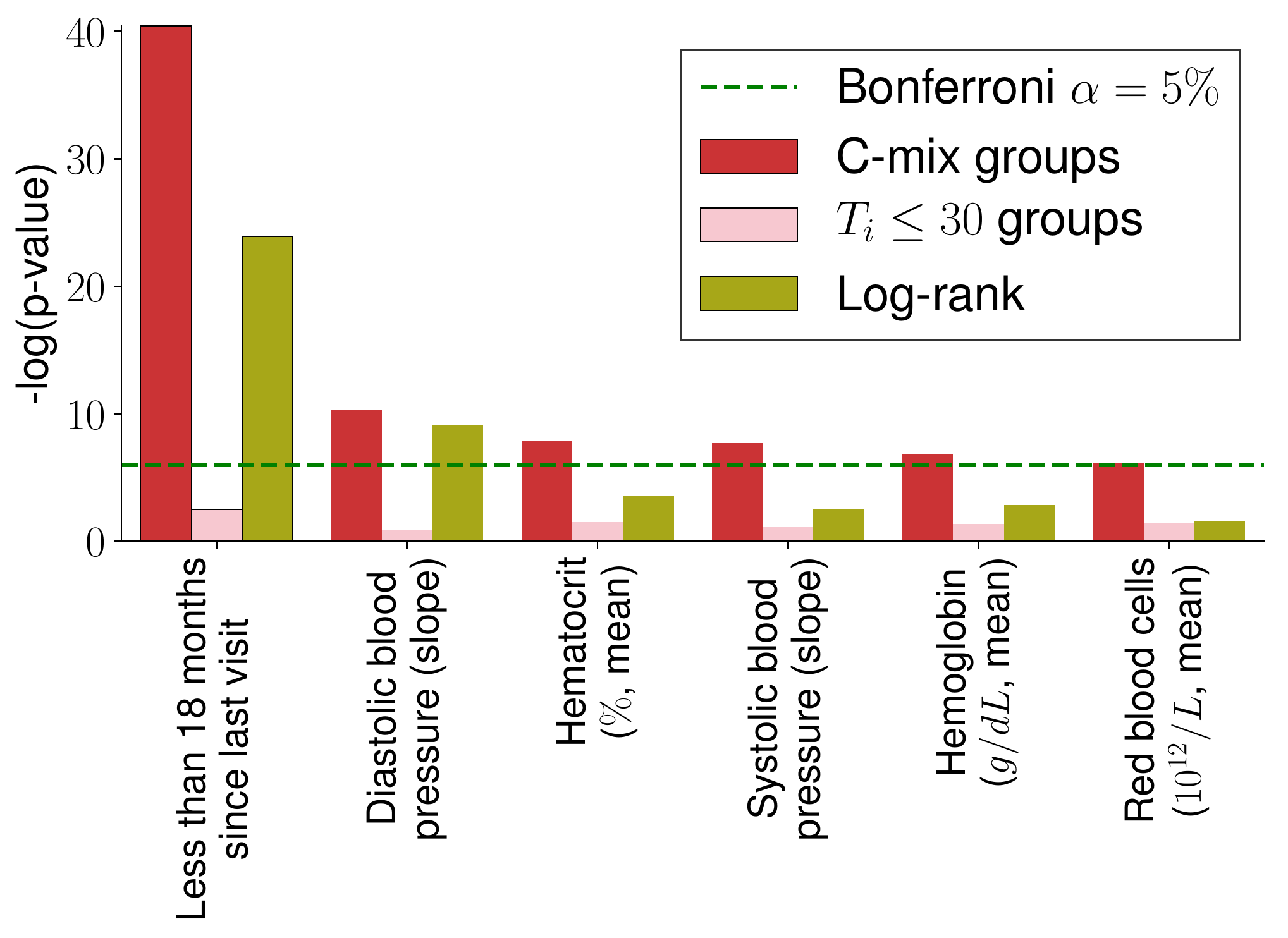}
\caption{Comparison of the tests based on the C-mix groups, on the $\epsilon=30$ days relative groups and on survival times. We arbitrarily shows only the tests with corresponding p-values below the level $\alpha=5\%$, with the classical Bonferroni multitests correction~\citep{bonferroni1935calcolo}.}
\label{fig:tests}
\end{figure}

Based on those early-readmission risk learned subgroups, we test for significant differences between them using Fisher-exact test~\citep{upton1992fisher} for binary covariate, and Wilcoxon rank-sum test~\citep{wilcoxon1945individual} for quantitative covariate. 
Then, we similarly test for significant difference, on each covariate, between naively created groups obtained with the binary outcome setting ($\epsilon=30$ days). We also use the log-rank test~\cite{harrington1982class} on each covariate, directly involving quantitative readmission delays. Finally, we compared the obtained significance (the p-value) for each test, on each covariate. The tests induced by the C-mix model are the most significant ones for almost all covariates.
The top-6 p-values of the tests are compared in Figure~\ref{fig:tests}. 
\begin{figure*}[]
\captionsetup{width=0.95\linewidth}
\centering
\includegraphics[width=.96\textwidth]{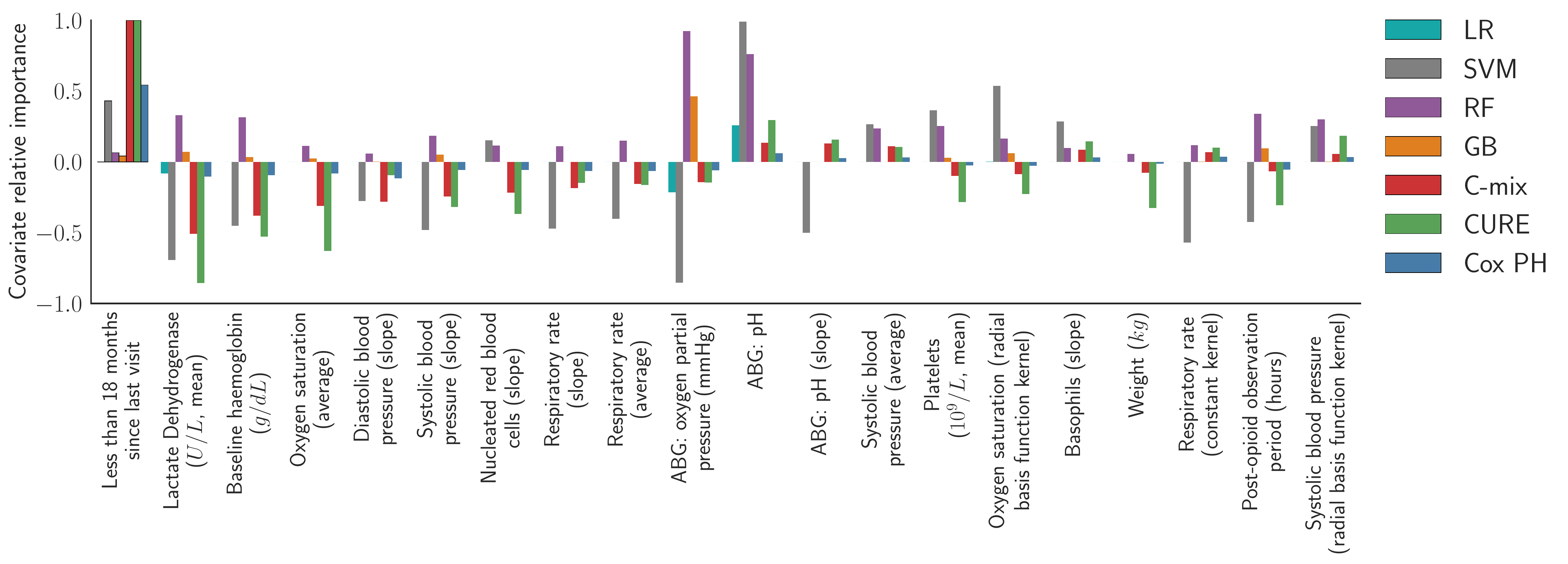}
\caption{Comparison of the top-20 covariates importance ordered on the C-mix estimates. Note that some time-dependent covariates, such as average cinetic during the last 48 hours of the stay (slope) or Gaussian Processes kernels parameters, appear to have significant importances.}
\label{fig:beta_20}
\end{figure*}

\begin{figure*}[!h]
\centering
\captionsetup{width=.96\linewidth}
\includegraphics[width=.96\textwidth]{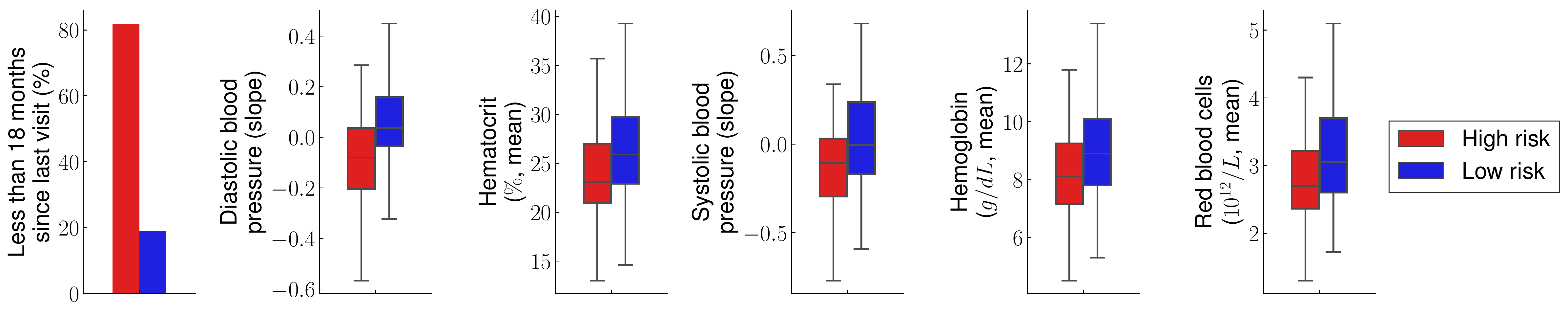}
\caption{Covariates boxplot comparison between the most significant C-mix groups.}
\label{fig:boxplot}
\end{figure*}

Taking the most significant C-mix groups highlighted in Figure~\ref{fig:tests}, Figure~\ref{fig:boxplot} shows either boxplot (for quantitative covariates) or repartition (for qualitative covariates) comparison between those groups. One can now easily visualize and interpret early-readmission risk data-driven grouping, and focus on specific covariate. For instance, it appears that patients among the high risk group tend to have a lower hemoglobin level, as well as a slightly lowering diastolic blood pressure in the last 48 hours of the stay (while slightly uppering for the low risk group). It also appears that less patients among the low risk group have visited the emergency department in the last 18 months.
\begin{figure}[!htb]
\captionsetup{width=0.85\linewidth}
\centering
\includegraphics[width=.95\linewidth]{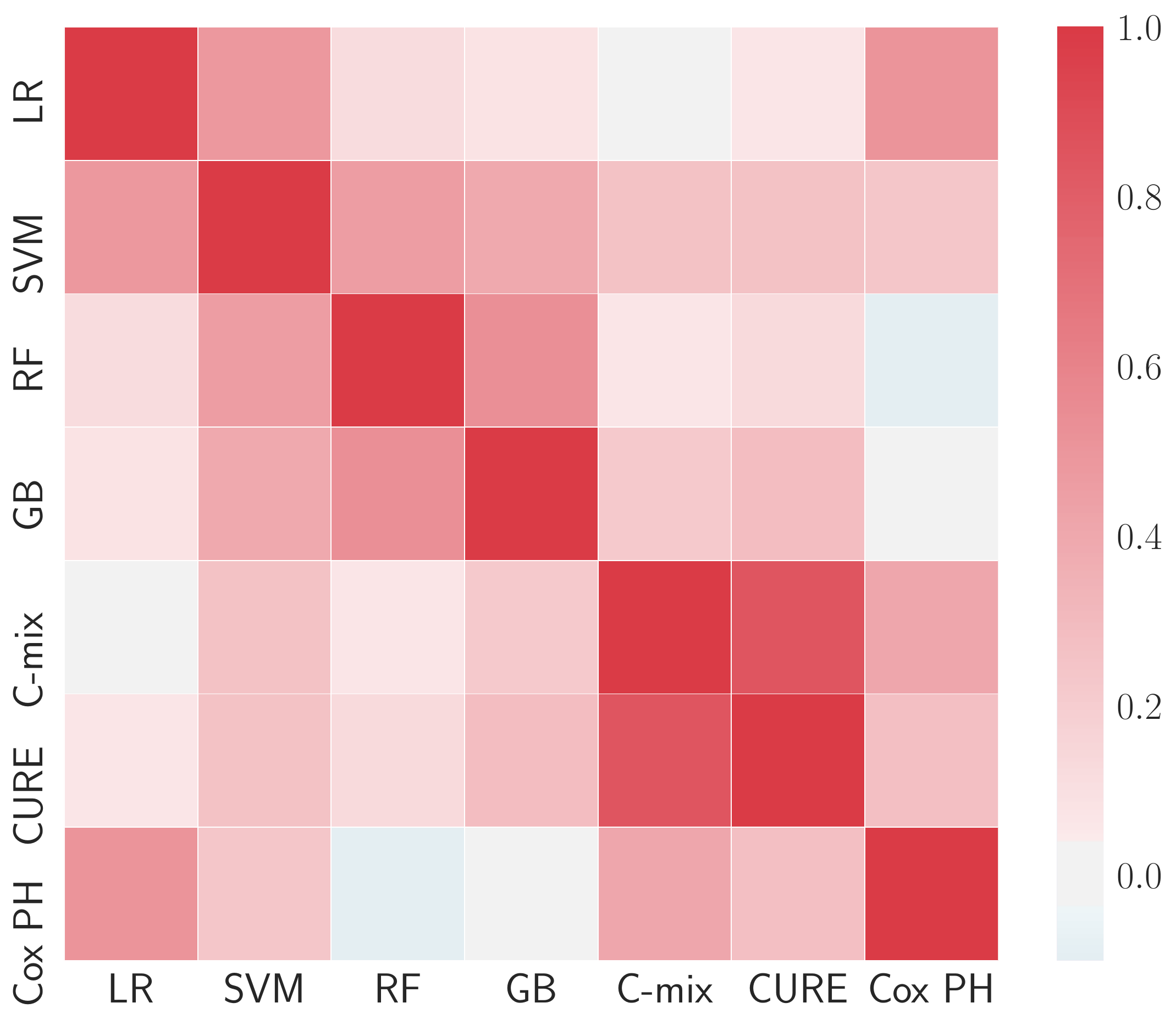}
\caption{Pearson correlation matrix for comparing covariates selection similarities between methods. Red means high correlations.}
\label{fig:corr}
\end{figure}

Let us now focus on the covariates selection aspect for each method.
Figure~\ref{fig:beta_20} gives an insight on the covariates importance relatively to each model for 20 arbitrarily chosen covariates (selected on decreasing importance order for the C-mix model). The result with all covariates can be found in Section~\ref{sec:beta_comparison} of Supplementary Material.
One can observe some consistency between methods. Figure~\ref{fig:corr} gives a global similarity comparison measure in terms of covariates selection. We observe higher similarities between methods within a single setting.

\section*{Discussion}

In this paper, rather than trying to be exhaustive in terms of considered methods, we choose, accordingly with the aim of this paper, to offer a methodology for fairly comparing methods in the two considered settings.
Also, we do not try different $\epsilon$ values, as it is done in~\citet{chen2012assessment} (where emphasis is on performance metrics), since our focus is to propose a general comparison and interpretation methodology, with an analysis that remains valid for any choice of $\epsilon$ value. 

In the binary outcome setting, classifiers highly depend on how the risk groups are defined: a slight change of the survival threshold $\epsilon$ for assignment of classes can lead to different prediction results~\citep{chen2012assessment}.
In our dataset, only 5.2\% of the visits lead to a readmission within 30 days. We are then in a classical setup where the adverse event appears rarely in the data at our disposal. 
In such setting, a vast amount of temporal information is lost since the model only knows if a readmission occurs before the threshold delay or not. 
It appears that taking all the information through the survival analysis setting first, and then going back to a binary prediction using the survival estimate, significantly enhances any binary prediction, which intuitively makes sense.

Among all methods, the C-mix holds the best results. Its good performances compared to other methods is already shown in~\citet{bussy2016}, both in synthetic and real data.
While the Cox PH regression model is widely used to analyze time-to-event data, it relies on the proportional hazard ratio assumption. 
But in the case of VOC for instance, it is plausible that these early-readmissions are the consequences of the same ongoing crisis (hospital discharge before the VOC recovery), whereas late-readmissions are genuine new unrelated crisis (recurrence). This would suggest that the proportional hazard ratio assumption for Cox PH model (or its related models like the competing risks model, the marginal model or the frailty model; for this reason not considered in this study) is not respected in this situation.
The CURE model main hypothesis being that a proportion of the patient is cured is questionable too.
Those reasons partly explain the good performances of the C-mix model that does not rely on any restrictive hypothesis. 

In this study, data extraction was performed with no \textit{a priori} on the relevance of each variable. For instance, we extracted all biological covariates that have been measured during a patient’s stay, without presuming of their importance on readmission risk. Selected variables in our use case are relevant from a clinical point of view, highlighting the capacity of regularization methods to pinpoint clinically relevant covariates.

The most important covariates in the survival setting are linked to the severity of the underlying SCD (\textit{e.g.} crisis frequency, baseline hemoglobin), while selected covariates in the binary outcome setting are more related to the crisis biological parameters (\textit{e.g.} arterial blood gas parameters). Some covariates appear to be selected in both settings (\textit{e.g.} mean lactate deshydrogenase). 
All selected covariates make sens from a clinical point of view, and the difference between the two settings seems to be related to the underlying hypotheses of each setting: as binary setting only takes information on early readmission, crisis related parameters are favored; meanwhile in the survival setting, parameters related to the severity of the underlying SCD are favored. This underlines why it is crucial, when working on prognosis analysis, to use several methods to get an insight of the most important covariates.

\section*{Conclusions}

In this paper, we compare methods in terms of prediction performances and covariates selection for different statistical and machine learning methods on a readmission framework with high dimensional EHR data. We particularly focus on comparing survival and binary outcome settings. 
Methods from both settings must be considered when working on a prognosis study. Indeed, important covariates are possibly different depending on the setting: for instance in our case study, we highlight important covariates linked either to the severity of the underlying SCD or to the severity of the crisis.

Not only do frequent readmissions affect SCD patients’ quality of life, they also impact hospitals’ organization and induce unnecessary costs. Our study lays the groundwork for the development of powerful methods which could help provide personalized care. Indeed, such early-readmission risk-predicting tools could help physicians decide whether or not a specific patient should be discharged of the hospital. Nevertheless, most selected covariates were derived from raw or unstructured extracted data, making it difficult to implement the proposed predictive models into routine clinical practice.

All results in the binary outcome setting rely on a critical readmission delay choice, which is a questionable - if not counterproductive - bias in readmission risk studies. 
Additionally, we point out the idea that learning in the survival setting, rather than directly from the binary outcome setting, and then making binary predictions through the estimated survival function for a given delay threshold can dramatically enhance performances. 

Finally, the C-mix model yields the better performances and can be an interesting alternative to more classical methods found in the medical literature to deal with prognosis studies in a high dimensional framework. Moreover, it provides powerful interpretations aspects that could be useful in both clinical research and daily practice (see Figure~\ref{fig:boxplot}). It would be interesting to generalize our conclusions to external datasets, which is the purpose of further investigations.

\section*{List of abbreviations}
AUC, Area Under the (ROC) Curve; CDW, Clinical Data Warehouse; CPOE, Computerized Provider Order Entry; EHR, Electronic Health Record; GB, gradient boosting; GPUH, George Pompidou University Hospital; LOINC, Logical Observation Identifiers Names and Codes; NN, artificial neural networks; PH, Proportional Hazards; RF, random forest; ROC, Receiver Operating Characteristic; SCD, Sickle-Cell Disease; SVM, Support Vector Machine; VOC, Vaso-Occlusive Crisis.

\begin{backmatter}

\section*{Competing interests}
The authors declare that they have no competing interests. This study received approval from the institutional review board from Georges Pompidou University Hospital (IRB 00001072 - project n$^\circ$ CDW$\_2014\_0008$) and the French data protection authority (CNIL - n$^\circ$ 1922081).

\section*{Author's contributions}
Data representation was done by SB, imputation by RV and cleaning by SB and RV.
All authors contributed to the design, analysis and writing of this manuscript, with a major contribution of SB.
Most of the underlying code was implemented by SB.
ASJ, AG and SG participated in the planning and supervision of the study.
All authors participated in the results interpretation.
All authors reviewed and revised the draft version of the manuscript. All authors read and approved the final version of the manuscript.

\section*{Acknowledgements}

We thank the reviewers for insightful comments that improved the presentation of the manuscript. We also thank Eric Zapletal for his help to extract the data.

\section*{Additional Files}

\subsection*{Additional file 1 --- Supplementary Material}

Supplementary material is given alongside the main manuscript, providing additional tables, figures or technical details mentionned in the manuscript. References to the right section of the Supplementary Material are precised throughout the paper and as soon as necessary. 

\subsection*{Additional file 2 --- C-mix software}

The C-mix (as well as CURE as a particular case) high-dimensional implementation is available online as an open-source project. We point this out since all other methods used in this manuscript are readily accessible in almost all development framework, which is not the case for the C-mix model.
Some details are given in Table~\ref{table:c-mix-code} below.
\begin{table}[!htb]
\centering
\caption{Details on the C-mix implementation. Tutorials are provided on the project page to make it easy to try the model on any data.}
\begin{tabular}{cc}
\toprule
Project name & C-mix code \\
Project home page & \href{https://github.com/SimonBussy/C-mix}{https://github.com/SimonBussy/C-mix} \\
Operating system(s) & Platform independent \\
Programming language & Python \\
License & MIT License \\
\bottomrule
\end{tabular}
\label{table:c-mix-code}
\end{table}

\bibliographystyle{plainnat} 
\bibliography{biblio} 

\end{backmatter}
\end{document}

\begin{frontmatter}

\begin{fmbox}
\dochead{Research}

\title{Supplementary Material for the paper: \\
Comparison of methods for early-readmission prediction in a high-dimensional heterogeneous covariates and time-to-event outcome framework}

\author[
   addressref={aff1},
   corref={aff1},
   email={simon.bussy@gmail.com}
]{\inits{SB}\fnm{Simon} \snm{Bussy}}
\author[
   addressref={aff2,aff3}        
]{\inits{RV}\fnm{Rapha\"el} \snm{Veil}}
\author[
   addressref={aff2,aff3}        
]{\inits{VL}\fnm{Vincent} \snm{Looten}}
\author[
   addressref={aff3}
]{\inits{AB}\fnm{Anita} \snm{Burgun}}
\author[
   addressref={aff1,aff4}        
]{\inits{SG}\fnm{St\'ephane} \snm{Ga\"iffas}}
\author[
   addressref={aff5}        
]{\inits{AG}\fnm{Agathe} \snm{Guilloux}}
\author[
   addressref={aff6,aff7}        
]{\inits{BR}\fnm{Brigitte} \snm{Ranque}}
\author[
   addressref={aff2,aff3}
]{\inits{ASJ}\fnm{Anne-Sophie} \snm{Jannot}}

\address[id=aff1]{
  \orgname{Laboratoire de Probabilités Statistique et Modélisation (LPSM), UMR 8001, Sorbonne University},
  \street{4 Place Jussieu},
  \postcode{75005}
  \city{Paris},
  \cny{France}
}
\address[id=aff2]{                   
  \orgname{Biomedical Informatics and Public Health Department, European Georges Pompidou Hospital, Assistance Publique-H\^opitaux de Paris}, 
  \street{20 Rue Leblanc},                     
  \postcode{75015}                                
  \city{Paris},                             
  \cny{France}                                    
}
\address[id=aff3]{
  \orgname{INSERM, UMRS 1138 team 22, Centre de Recherche des Cordeliers, Université Paris Descartes},
  \street{15 Rue de l'École de Médecine},
  \postcode{75006}
  \city{Paris},
  \cny{France}
}
\address[id=aff4]{
  \orgname{CMAP, UMR 7641 École Polytechnique CNRS},
  \street{Route de Saclay},
  \postcode{91128}
  \city{Palaiseau},
  \cny{France}
}
\address[id=aff5]{
  \orgname{LAMME, Univ Evry, CNRS, Universit\'e Paris-Saclay, France},
  \street{23 boulevard de France},
  \postcode{91025}
  \city{Evry},
  \cny{France}
}
\address[id=aff6]{
  \orgname{INSERM UMRS 970, Université Paris Descartes},
  \street{56 rue Leblanc},
  \postcode{75015}
  \city{Paris},
  \cny{France}
}
\address[id=aff7]{
  \orgname{Assistance Publique-Hôpitaux de Paris, Internal Medicine Department, Georges Pompidou European Hospital},
  \street{20 Rue Leblanc},
  \postcode{75015}
  \city{Paris},
  \cny{France}
}

\end{fmbox}

\end{frontmatter}

\section{Details on covariates}
\label{sec:covariates-details}

\subsection{Covariates creation}
\label{sec:cov_creation}

Since SCD patients are frequently treated with opioids to control the pain induced from VOCs, some may develop, over time, an addiction to these products. Such addiction may cause readmission and often interferes with hospitalization timeline. In order to limit confusion bias, we excluded patients encoded as opioid addicts (ICD-10 F11) as well as those who were treated with substitute products such as Methadone or Buprenorphine, both determined from hospitalization reports and drug prescriptions.

Regarding opioid treatment related information from the CDW, based on doctors and nurses inputs, variables extracted were the following: \\
- the specific molecule of each prescription, \\
- the specific dosage form of each prescription, \\
- the initiation and ending timestamps of each prescription. \\
From these variables, we also derived the following: \\
- the delay between the end of the last syringe received and the hospital discharge, \\
- the number of syringes used per day on average, \\
- the slope from the linear regression of the delay between syringes throughout the stay.

Regarding intravenous opioid treatments, we also extracted bolus dosage, maximum dosage, and refractory period. In order to capture both the average level and the general trend of these covariates, we derived them by calculating the slope and intercept from the linear regression of each of these parameters throughout the stay.

\subsection{Missing data}
\label{sec:missing_data}

We substitute missing medical history related data as follows: if a specific medical condition or VOC complication is mentioned in a report, this item is considered as part of the patient' medical history for every chronologically following stays; if a specific medical condition or VOC complication is explicitly stated as absent from the medical history in a report, this item is considered absent in all the previous stays.\\
For other specific covariates, we proceed that way:\\
- for the patients’ baseline hemoglobin value, we use the last hemoglobin value measured during the first included stay, \\
- for the dichotomous variables regarding the patient’s entourage and professional activity, we use the most represented value amongst all stays (of all patients), \\
- we consider non-mentioned medical history or VOC complications as absent, \\
- we consider that all patients received both opioid treatments and oxygen therapy at admission in the emergency room. Therefore, we consider the post-opioid observation period, as well as the post-oxygen observation period, to be the same time length as the entire stay.\\
For all remaining covariates, we impute as follows (after the random sampling of one stay per patient):\\
- numerical variables are imputed with their median values,\\
- categorical variables are imputed with their most represented values.

\subsection{List of covariates}
\label{sec:covariates_list}

Table~\ref{table:covariates} summarizes the concepts used and their basic properties.

\begin{sidewaystable}[!htb]
\centering
\caption{List of the considered concepts. For each one, we display the name (with unit), the category, the sub-category if relevant, and the type (``Q'' for Qualitative, ``B'' for Binary and ``C'' for Categorical). For practical purposes, we only display basic concepts without describing the list of covariates induced from it and used in practice, since the process of covariates extraction is thoroughly
described in the paper. For instance, the temperature concept gives rise to 5 covariates, relatively to its average and slope in the last 48 hours as well as the corresponding Gaussian Process kernel hyper-parameters.}
\resizebox{\textwidth}{!}{
\begin{tabular}{ccccc|ccccc}
\toprule
Name (unit) & Category & Sub-category & Type & \hspace{.2cm} & \hspace{.2cm} & Name (unit) & Category & Type \\
\midrule
Red blood cells ($10^{12}/L$) & Biological data & Complete blood count  & Q & \hspace{.2cm} & \hspace{.2cm} & Respiratory rate (mvt/min) & Clinical data & Q \\
Hemoglobin ($g/dL$) & Biological data & Complete blood count  & Q & \hspace{.2cm} & \hspace{.2cm} & Heart rate (bpm) & Clinical data & Q \\
Haemoglobin gap to baseline ($g/dL$)  & Biological data & Complete blood count  & Q & \hspace{.2cm} & \hspace{.2cm} & Oxygen saturation ($\%$)  & Clinical data & Q \\
Hematocrit ($\%$) & Biological data & Complete blood count  & Q & \hspace{.2cm} & \hspace{.2cm} & Temperature ($^\circ C$) & Clinical data & Q \\
Mean cell volume ($fl$) & Biological data & Complete blood count  & Q & \hspace{.2cm} & \hspace{.2cm} & Post-oxygen observation period (hours)  & Clinical data & Q \\
Mean corpuscular hemoglobin ($pg$)  & Biological data & Complete blood count  & Q & \hspace{.2cm} & \hspace{.2cm} & Systolic blood pressure ($mmHg$) & Clinical data & Q \\
Mean corpuscular hemoglobin concentration ($\%$)  & Biological data & Complete blood count  & Q & \hspace{.2cm} & \hspace{.2cm} & Diastolic blood pressure ($mmHg$) & Clinical data & Q \\
Reticulocytes ($10^9/L$)  & Biological data & Complete blood count  & Q & \hspace{.2cm} & \hspace{.2cm} & Gender  & General features  & B \\
Nucleated red blood cells ($10^9/L$)  & Biological data & Complete blood count  & Q & \hspace{.2cm} & \hspace{.2cm} & Baseline haemoglobin ($g/dL$) & General features  & Q \\
White blood cells ($10^9/L$)  & Biological data & Complete blood count  & Q & \hspace{.2cm} & \hspace{.2cm} & Genotype  & General features  & B \\
Neutrophils ($10^9/L$)  & Biological data & Complete blood count  & Q & \hspace{.2cm} & \hspace{.2cm} & Distance between home and GPUH ($km$) & General features  & Q \\
Neutrophils ($\%$)  & Biological data & Complete blood count  & Q & \hspace{.2cm} & \hspace{.2cm} & Driving time from home to GPUH (minutes)  & General features  & Q \\
Basophils ($10^9/L$)  & Biological data & Complete blood count  & Q & \hspace{.2cm} & \hspace{.2cm} & Age at hospital admission & General features  & Q \\
Basophils ($\%$)  & Biological data & Complete blood count  & Q & \hspace{.2cm} & \hspace{.2cm} & French DRG code (GHM) & General features  & C \\
Eosinophils ($10^9/L$)  & Biological data & Complete blood count  & Q & \hspace{.2cm} & \hspace{.2cm} & Severity level of the stay  & General features  & C \\
Eosinophils ($\%$)  & Biological data & Complete blood count  & Q & \hspace{.2cm} & \hspace{.2cm} & Length of hospital stay (hours) & General features  & Q \\
Monocytes ($10^9/L$)  & Biological data & Complete blood count  & Q & \hspace{.2cm} & \hspace{.2cm} & Time length since last admission (days) & General features  & Q \\
Monocytes ($\%$)  & Biological data & Complete blood count  & Q & \hspace{.2cm} & \hspace{.2cm} & Less than 18 months since last admission  & General features  & Q \\
Lymphocytes ($10^9/L$)  & Biological data & Complete blood count  & Q & \hspace{.2cm} & \hspace{.2cm} & Time length to next admission (days)  & General features  & Q \\
Lymphocytes ($\%$)  & Biological data & Complete blood count  & Q & \hspace{.2cm} & \hspace{.2cm} & Stayed in ICU & General features  &  B \\
Platelets ($10^9/L$)  & Biological data & Complete blood count  & Q & \hspace{.2cm} & \hspace{.2cm} & Number of RBC transfusions  & General features  & Q \\
Mean platelet volume ($fl$) & Biological data & Complete blood count  & Q & \hspace{.2cm} & \hspace{.2cm} & Professional activity & Lifestyle & B \\
Hemoglobin S ($\%$) & Biological data & Hemoglobin electrophoresis  & Q & \hspace{.2cm} & \hspace{.2cm} & Household situation & Lifestyle & B \\
Hemoglobin F ($\%$) & Biological data & Hemoglobin electrophoresis  & Q & \hspace{.2cm} & \hspace{.2cm} & Acute chest syndrom & Medical history & B \\
Asparate transaminase ($U/L$) & Biological data & Liver function test & Q & \hspace{.2cm} & \hspace{.2cm} & Avascular bone necrosis & Medical history & B \\
Alanine transaminase ($U/L$)  & Biological data & Liver function test & Q & \hspace{.2cm} & \hspace{.2cm} & Priapism (only for males) & Medical history & B \\
Alkaline phosphatase ($U/L$)  & Biological data & Liver function test & Q & \hspace{.2cm} & \hspace{.2cm} & Ischemic stroke & Medical history & B \\
Gamma glutamyl-tranferase ($U/L$) & Biological data & Liver function test & Q & \hspace{.2cm} & \hspace{.2cm} & Leg skin ulceration & Medical history & B \\
Direct bilirubin ($mol/L$)  & Biological data & Liver function test & Q & \hspace{.2cm} & \hspace{.2cm} & Heart failure & Medical history & B \\
Total bilirubin ($mol/L$) & Biological data & Liver function test & Q & \hspace{.2cm} & \hspace{.2cm} & Pulmonary hypertension  & Medical history & B \\
Urea ($mmol/L$) & Biological data & Renal function test & Q & \hspace{.2cm} & \hspace{.2cm} & Known nephropathy & Medical history & B \\
Creatinine ($mol/L$)  & Biological data & Renal function test & Q & \hspace{.2cm} & \hspace{.2cm} & Known retinopathy & Medical history & B \\
Renal function by MDRD ($mL/min/1,73m^2$) & Biological data & Renal function test & Q & \hspace{.2cm} & \hspace{.2cm} & Dialysis  & Medical history & B \\
Sodium ($mmol/L$) & Biological data & Serum electrolytes  & Q & \hspace{.2cm} & \hspace{.2cm} & Received Morphine & Opioid use  & B \\
Potassium ($mmol/L$)  & Biological data & Serum electrolytes  & Q & \hspace{.2cm} & \hspace{.2cm} & Received Oxycodone  & Opioid use  & B \\
Chloride ($mmol/L$) & Biological data & Serum electrolytes  & Q & \hspace{.2cm} & \hspace{.2cm} & Received orally administered opioids  & Opioid use  & B \\
Bicarbonate ($mmol/L$)  & Biological data & Serum electrolytes  & Q & \hspace{.2cm} & \hspace{.2cm} & Number of syringes received per day & Opioid use  & Q \\
Total calcium ($mmol/L$)  & Biological data & Serum electrolytes  & Q & \hspace{.2cm} & \hspace{.2cm} & Delay between syringes (slope)  & Opioid use  & Q \\
Proteins ($g/L$)  & Biological data & Serum electrolytes  & Q & \hspace{.2cm} & \hspace{.2cm} & Post-opioid observation period (hours)  & Opioid use  & Q \\
Glucose ($mmol/L$)  & Biological data & Serum electrolytes  & Q & \hspace{.2cm} & \hspace{.2cm} & Bolus dosage  & Opioid use  & Q \\
C-reactive protein ($mg/L$) & Biological data & Other blood markers & Q & \hspace{.2cm} & \hspace{.2cm} & Maximum dosage & Opioid use  & Q \\
Lactate Dehydrogenase ($U/L$) & Biological data & Other blood markers & Q & \hspace{.2cm} & \hspace{.2cm} & Refractory period & Opioid use  & Q \\
Weight ($kg$) & Clinical data & Body dimensions & Q  \\
Size ($cm$) & Clinical data & Body dimensions & Q  \\
Body mass index ($kg/m^2$)  & Clinical data & Body dimensions & Q \\
\bottomrule
\end{tabular}}
\label{table:covariates}
\end{sidewaystable}

\section{Details on experiments}
\label{sec:details-results}

\subsection{Survival function estimation}
\label{sec:survival_estimation}

For the Cox PH model, the survival $\bP[T_i > t|X_i=x_i]$ for patient $i$ in the test set is estimated by
\begin{equation*}
\hat S_i(t|X_i=x_i) = [\hat S^\text{cox}_0(t)]^{\exp(x_i^\top \hat{\beta})},
\end{equation*}
where $\hat S_0^\text{cox}$ is the estimated survival function of baseline population ($x = 0$) obtained  using the Breslow estimate of $\lambda_0$ \citep{breslow1972contribution}.
For the CURE or the C-mix models, it is naturally estimated by 
\begin{equation*}
\hat S_i(t|X_i=x_i) = \pi_{\hat{\beta}}(x_i) \hat S_1(t) + \big(1 - \pi_{\hat{\beta}}(x_i) \big) \hat S_0(t),
\end{equation*} 
where $\hat S_0$ and $\hat S_1$ are the Kaplan-Meier estimators \citep{kaplan1958nonparametric} of the low and high risk of early-readmission subgroups respectively learned by the C-mix model (patients with $\pi_{\hat{\beta}}(x_i) > 0.5$ are clustered in the high risk subgroup, others in the low risk one), or cured and uncured subgroups respectively learned by the CURE model.

\subsection{Hyper-parameters tuning}
\label{sec:hyper_parameters}

Let us summarize the hyper-parameters obtained after the cross-validation procedure for each method. First, we take $\eta = 0.1$ for all method using Elastic-Net regularization to ensure covariates selection. The strengh of the penalty is tuned to 42.81 for LR, 0.05 for SVM, 0.03 for C-mix, 0.008 for CURE and 0.014 for Cox PH. For RF, the maximum depth is 7, the minium sample's split is 3, the minimum sample's leaf is 2, the criterion is the entropy and the number of estimators is tuned to 200. For GB, the maximum depth is 7, the minimum sample's split is 3, the minimum sample's leaf is 4 and the number of estimators is 200. Finally for NN, the hidden layer's sizes is 3, the regularization term is tuned to 0.13.

\subsection{Covariates importance comparison}
\label{sec:beta_comparison}

\begin{figure*}[!htb]
\captionsetup{width=0.85\linewidth}
\centering
\begin{subfigure}{.3\textwidth}
  \centering
  \includegraphics[width=.85\linewidth]{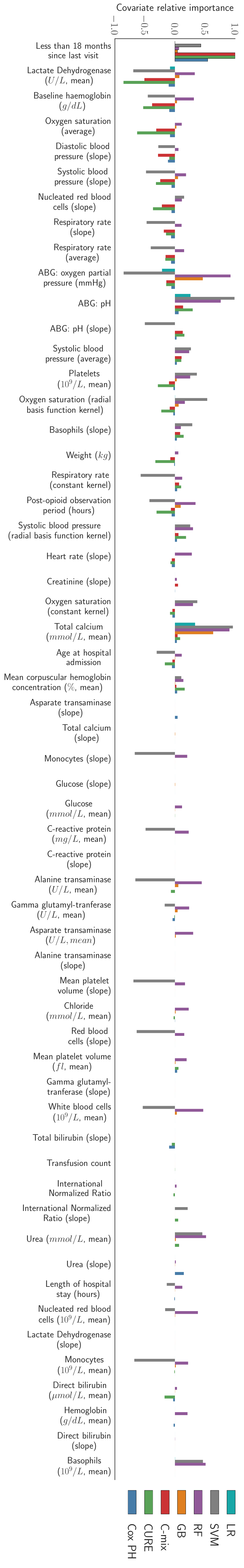}
\end{subfigure}%
\begin{subfigure}{.3\textwidth}
  \centering
  \includegraphics[width=.85\linewidth]{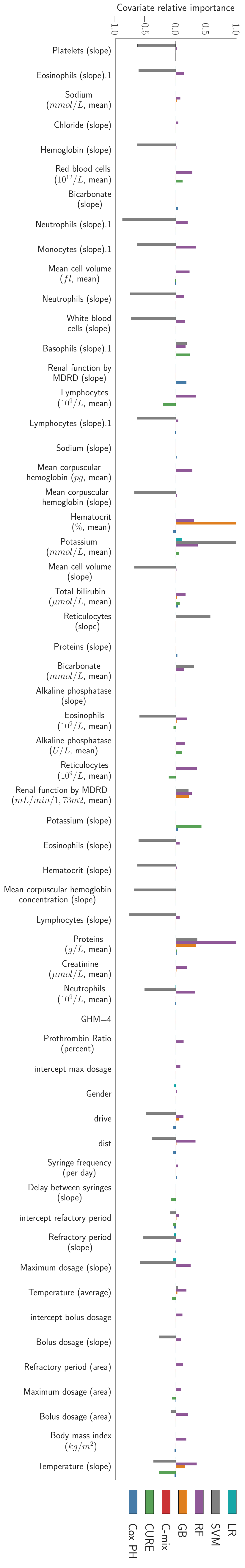}
\end{subfigure}
\begin{subfigure}{.3\textwidth}
  \centering
  \includegraphics[width=.85\linewidth]{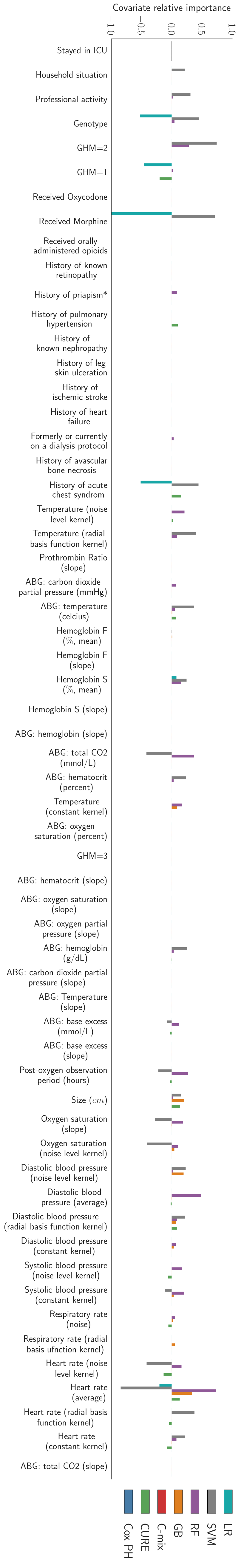}
\end{subfigure}
\caption{Comparison of covariates importance, ordered on the C-mix estimates. Note that for RF and GB models, coefficients are, by construction, always positive.}
\label{fig:coeff_comparison}
\end{figure*}

Figure~\ref{fig:coeff_comparison} gives the covariates importance estimates for all covariates and all considered methods.

\begin{backmatter}

\bibliographystyle{plainnat} 
\bibliography{biblio} 

\end{backmatter}